\documentclass[runningheads]{llncs}
\usepackage{graphicx}
\usepackage{listings}
\usepackage[ruled,linesnumbered,vlined]{algorithm2e}
\usepackage{xspace}

\usepackage{amsmath}
\usepackage{amssymb}
\usepackage{color}

\usepackage{listings}
\makeatletter
\lst@Key{countblanklines}{true}[t]{\lstKV@SetIf{#1}\lst@ifcountblanklines}
\lst@AddToHook{OnEmptyLine}{%
    \vspace{-0.7\baselineskip}%
    \lst@ifnumberblanklines\else%
    \lst@ifcountblanklines\else%
    \advance\c@lstnumber-\@ne\relax%
    \fi%
    \fi}
\makeatother
\newcommand\derives{\ensuremath{\scalebox{0.6}[1.2]{\(:\!- \)}}}

\newcommand{\tuple}[1]{\ensuremath{\langle{#1}\rangle}\xspace}
\newcommand{\naf}[0]{\mathit{not}\phantom{x}}
\newcommand{\malvi}[1]{\textcolor[rgb]{0.25,0.5,0.25}{#1}}

\newcommand{\gibbi}[1]{\textcolor{blue}{#1}}

\newcommand{\dquo}[1]{``#1"}

\newcommand{\nop}[1]{}

\begin{document}
\title{Rethinking Answer Set Programming Templates}%\thanks{Supported by organization x.}}
\author{Mario Alviano\orcidID{0000-0002-2052-2063} \and
Giovambattista Ianni\orcidID{0000-0003-0534-6425} \and
Francesco Pacenza\orcidID{0000-0001-6632-3492} \and
Jessica Zangari\orcidID{0000-0002-6418-7711}
}
\authorrunning{M. Alviano et al.}
\institute{DEMACS, University of Calabria, Via Bucci 30/B, 87036 Rende (CS), Italy\\
\email{\{name.surname\}@unical.it}}
\maketitle              % typeset the header of the contribution
\begin{abstract}
In imperative programming, the Domain-Driven Design methodology helps in coping with the complexity of software development by materializing in code the invariants of a domain of interest.
Code is cleaner and more secure because any implicit assumption is removed in favor of invariants, thus enabling a fail fast mindset and the immediate reporting of unexpected conditions.
This article introduces a notion of template for Answer Set Programming that, in addition to the {\em don't repeat yourself} principle, enforces locality of some predicates by means of a simple naming convention.
Local predicates are mapped to the usual global namespace adopted by mainstream engines, using universally unique identifiers to avoid name clashes. This way, local predicates can be used to enforce invariants on the expected outcome of a template in a possibly empty context of application, independently by other rules that can be added to such a context.
Template applications transpiled this way can be processed by mainstream engines and safely shared with other knowledge designers, even when they have zero knowledge of templates.


\end{abstract}

\section{Introduction}\label{sec:intro}

Answer Set Programming (ASP) is a declarative specification language suited to address combinatorial search and optimization \cite{DBLP:journals/aim/ErdemGL16,DBLP:books/sp/Lifschitz19}.
In ASP, problem requirements are represented in terms of a program made of logic rules, and solutions are obtained by computing stable models of the program, that is, classical models satisfying an additional stability condition.
Fast prototyping is very likely the most appreciated strength of ASP, as it provides several linguistic constructs to ease the representation of complex knowledge and allows for quickly testing alternative solutions to the same problem of interest \cite{DBLP:conf/jelia/AmendolaBR21}.
The linguistic capabilities of ASP are accompanied by several efficient algorithms addressing a broad variety of computational problems, therefore different solving strategies can be attempted with minimal design effort \cite{DBLP:journals/tplp/SonPBS23}.
Nonetheless, ASP has a few shortcomings when used in broad development environments. %, which are detailed below.
First of all, ASP programs are often seen as a whole, with rules interacting in any possible way.
This is in part due to the stable model semantics adopted by ASP, which is nonmonotonic and therefore not friendly to the definition of \emph{invariants}~\cite{DBLP:journals/tplp/FandinnoLLS20,DBLP:journals/corr/abs-1905-03196}.
An invariant, literally \emph{something that does not change or vary}, is an assumption taken by a block of code to guarantee correctness of computation.
For example, consider a function written in C to implement the factorial of a number $n$ stored as \lstinline|uint32_t| by multiplying all positive integers less than or equal to $n$.
Such a function is correct under the assumptions that $n \leq 12$, as $13!$ exceeds the limit of \lstinline|uint32_t| (and an integer overflow would occur).
If the function starts by raising an error when $n > 12$, integer overflows are impossible and the computation of the factorial is guaranteed to be correct.
In ASP, guaranteeing that some properties of the stable models of a program are preserved when the program is extended with other rules is nontrivial~\cite{DBLP:journals/tocl/BomansonJN20,DBLP:journals/tplp/Fink11,DBLP:journals/tocl/OetschSTW21}.
On the other hand, invariants greatly help programmers to reason about their code, for example when looking for bugs, and therefore the difficulty to introduce invariants in ASP programs should not lead to underestimating their potential benefits.

Ideally, a programming language should have linguistic constructs to ease the reuse of code \cite{DBLP:journals/isse/AlOmarWRMNO22}.
Macros, subroutines and templates are often used to address such a concern, and some constructs in this direction were defined and implemented also for ASP~\cite{DBLP:conf/iclp/BaralDT06,DBLP:journals/aicom/CalimeriI06,DBLP:conf/kr/Dao-TranEFK10}.
Much work suggests that some predicates should be considered \emph{hidden}, essentially \emph{auxiliary} to the definition of other \emph{visible} predicates that actually define the semantics of programs or modules of a program \cite{DBLP:journals/tocl/BomansonJN20,DBLP:journals/tplp/CabalarFL20}.
As such, auxiliary predicates should not be taken into account when checking equivalence of programs, and should not be shared between different contexts.
The formalization of such an intuition is not necessarily trivial, and of course the complexity of its implementation and successful adoption by practitioners strongly depend on how easy is to specify that a predicate is auxiliary.
This work contributes to the reuse of ASP code by introducing and implementing a simple notion of \emph{template}. 
The notion of template is here intended in the broad sense of a set of rules, not explicitly identifying input and output predicates, with the capability of being applied multiple times by specifying possibly different renaming mappings for \emph{visible} predicates.
Auxiliary predicates are kept \emph{local} to their template applications by an automatic renaming policy that appends a \emph{universally unique identifier} (UUID~\cite{RFC4122}), generated at application time, to predicate names.
This way, visible and auxiliary predicates of different template applications can coexist in the usual \emph{global} namespace adopted by ASP engines.
To ease the implementation of this idea and its adoption by practitioners, local predicates are automatically identified by introducing a naming convention commonly used in Python objects to declare \emph{private} attributes and methods, that is, names starting by \emph{double underscore} are associated with local predicates.
%
% STATIC: REMOVED FOR NOW
% In addition to local variables, whose scope and lifetime is restricted to the function they are declared in, some programming languages support \emph{static} local variables.
% The scope of a static local variable is still restricted to the function it is declared in, but its lifetime is extended to the lifetime of the program.
% In ASP the notion of static local variable can be implemented by enriching the naming convention for local variables:
% names starting by \lstinline|__static_| in a template are renamed by appending a UUID generated at template declaration time, so that the new name is shared among different applications of the same template, still avoiding any naming collision between different templates.

Templates of a complex system should be \emph{tested as a unit} before looking at their behavior in the integrated final program.  
Following the \emph{Test-Driven Development} (TDD) methodology, requirements emerging during the analysis of a domain of interest are made explicit in code by means of tests, and business code is then implemented with the goal of passing tests;
usually, one requirement at time is addressed, and tests are run every time the business code is modified, either because of a new requirement is fulfilled, or because a code refactoring is needed.
In order to enable the verification of some invariants that a template imposes on any program it is applied to, we focus on \emph{here-and-there} models, that is, models of the monotonic, modal logic of here-and-there.
Intuitively, here-and-there models of a program $\Pi$ are pairs of the form $\tuple{H,T}$ such that $T$ is a stable model candidate, and $H$ is a model of the program reduct $\Pi^I$, so that stable models are characterized as here-and-there models of the form $\tuple{T,T}$ such that there is no here-and-there model $\tuple{H,T}$ with $H \subset T$ \cite{DBLP:journals/algorithms/FandinnoPVW22,DBLP:journals/amai/Pearce06}.
Being a monotonic logic, here-and-there models of a set of rules are necessarily an overestimate of the here-and-there models of any broader set of rules.
This is an invariant that can power the definition of tests providing guarantees on the behavior of a template;
for example, it is possible to define tests to guarantee that some atoms have a specific assignment in all stable models of any program extending a template application, that some interpretations cannot be stable models even if consistent with all rules of a broader program, or also to guarantee coherence of some atoms w.r.t.\ some interpretations (intended as derivation in the associated program reducts).

In summary, the contributions of this work are the following:
\begin{itemize}
\item 
We propose a notion of template made of a set of rules with global and local predicates.
Templates can be applied and mixed with other programs by means of a versatile form of predicate renaming that overcomes the inflexible practice of fixing input and output predicates, and therefore broadens the reusability of code snippets.
Global predicates are mapped according to the preferences of knowledge designers, while the renaming of local predicates is automatic and clash-free without the need for synchronizing template applications.

\item 
We enable the possibility to enforce some invariants of a template, some of which can be verified by analyzing its here-and-there models with the help of mainstream ASP engines.
A context of application for the template is possibly given by specifying other rules, and invariants are obtained thanks to the impossibility to refer local predicates outside the template.

\item 
We define a transpiler that expands programs with templates to ordinary ASP programs.
Transpiled programs can be evaluated by mainstream ASP engines, and combined with other programs with a practical guarantee that local predicate names do not clash.
Our transpiler and its testing constructs are powered by the \textsc{clingo python api}.
\end{itemize}

\section{Background}\label{sec:background}

A \emph{universally unique identifier} (UUID) is a 128-bit label generated according to standard methods that guarantee uniqueness for practical purposes \cite{RFC4122} (i.e., even if the probability of generating duplicated UUIDs is not zero, it is generally considered close enough to zero to be negligible).

A \emph{normal program} is a set of rules of the form
\begin{equation}\label{eq:rule}
    p_0(\overline{t_0}) \leftarrow p_1(\overline{t_1}),\ \ldots,\ p_m(\overline{t_m}),\ \naf p_{m+1}(\overline{t_{m+1}}),\ \ldots,\ \naf p_n(\overline{t_n})
\end{equation}
where $n \geq m \geq 0$, and each $p_i(\overline{t_i})$ is an atom made of a predicate $p_i$ from a fixed set $\mathbf{P}$ and a sequence $\overline{t_i}$ of terms;
terms are either variables (uppercase strings) from a fixed set $\mathbf{V}$ or constants (integers and lowercase strings) from a fixed set $\mathbf{C}$.
Sets $\mathbf{P}$, $\mathbf{V}$ and $\mathbf{C}$ are countably infinite and pairwise disjoint.
For a rule $r$ of the form \eqref{eq:rule}, let $H(r)$ denote the \emph{head} atom $p_0(\overline{t_0})$, and $B^+(r), B^-(r)$ denote the sets $\{p_i(\overline{t_i}) \mid i = 1..m\}$ and $\{p_i(\overline{t_i}) \mid i = m+1..n\}$ of atoms occurring in \emph{positive} and \emph{negative literals} of $r$.
A rule of the form \eqref{eq:rule} is a \emph{fact} if $n = 0$. 
We adopt the usual shortcut $p/n$ for referring to predicate $p$ of arity $n$.
Let $\mathit{pred}(\Pi)$ be the set of predicate names occurring in program $\Pi$. 

The \emph{grounding} $\mathit{grd}(\Pi)$ of $\Pi$ is $\bigcup_{r \in \Pi}{\mathit{grd}(r)}$, where $\mathit{grd}(r)$ is obtained from $r$ by replacing variables from $\mathbf{V}$ with constants from $\mathbf{C}$, in all possible ways.
An \emph{interpretation} $I$ is a set of ground atoms (i.e., atoms without variables):
atoms in $I$ are true, other atoms are false.
The relation $\models$ (is model of) is defined inductively:
for a ground atom $p(\overline{c})$, $I \models p(\overline{c})$ if $p(\overline{c}) \in I$, and $I \models \naf p(\overline{c})$ if $p(\overline{c}) \notin I$;
for a ground rule $r$, $I \models B(r)$ if $I$ is a model of all literals in $B(r)$, and $I \models r$ if $I \models H(r)$ whenever $I \models B(r)$;
for a program $\Pi$, $I \models \Pi$ if $I \models r$ for all $r \in \mathit{grd}(\Pi)$.
The \emph{reduct} $\Pi^I$ of $\Pi$ w.r.t.\ $I$ is %the ground program 
$\{H(r) \leftarrow B^+(r) \mid r \in \mathit{grd}(\Pi),$ $I \models B(r)\}$.
$I$ is a \emph{stable model} of $\Pi$ if $I \models \Pi$ and there is no $J \subset I$ such that $J \models \Pi^I$.
Let $\mathit{SM}(\Pi)$ be the set of stable models of $\Pi$.

\begin{example}\label{ex:asp}
Consider the following example in ASP-Core-2 syntax \cite{DBLP:journals/tplp/CalimeriFGIKKLM20}:
\begin{lstlisting}
    a(X) $\derives$ e(X), not b(X).      e(1).  e(2).  
    b(X) $\derives$ e(X), not a(X).      fail $\derives$ a(1), b(2), not fail.
\end{lstlisting}
The above program has three stable models, namely
$X = \{e(1),e(2)\}$, $X \cup \{a(1),a(2)\}$, $X \cup \{b(1),a(2)\}$, and $X \cup \{b(1),b(2)\}$.
It is using the fact that \lstinline|fail| only occurs in the head of rules of the form $\mathit{fail} \leftarrow \mathit{body},\ \naf \mathit{fail}$, a well-known pattern to simulate constraints in ASP.
\hfill$\blacksquare$
\end{example}
In the following, $\bot \leftarrow \mathit{body}$ is used as syntactic sugar for $\mathit{fail} \leftarrow \mathit{body},\ \naf \mathit{fail}$, where $\mathit{fail}$ is not used elsewhere (note that this assumption will be turned into an invariant by Example~\ref{ex:unstable}).

A program can be mapped to a theory of the logic of here-and-there (HT for short; \cite{HeytingDieFR}) by replacing each rule of the form \eqref{eq:rule} with a formula
\begin{align}\label{eq:ht-rule}
    p_1(\overline{t_1}) \wedge \cdots \wedge p_m(\overline{t_m}) \wedge (p_{m+1}(\overline{t_{m+1}}) \rightarrow \bot) \cdots \wedge (p_n(\overline{t_n}) \rightarrow \bot) \rightarrow p_0(\overline{t_0}) 
\end{align}
and by expanding variables with constants from $\mathbf{C}$ (as done for the grounding).
Let $\Gamma_\Pi$ denote the theory associated with program $\Pi$.
A HT-interpretation is a pair $\tuple{I_H,I_T}$ of interpretations such that $I_H \subseteq I_T$;
intuitively, $\tuple{I_H,I_T}$ represents two worlds, namely $H$ and $T$, with $H \leq T$.
Relation $\models$ is extended to $\tuple{I_H,I_T}$ and world $w \in \{H,T\}$ as follows:
$\tuple{I_H,I_T,w} \not\models \bot$;
for a ground atom $p(\overline{c})$, $\tuple{I_H,I_T,w} \models p(\overline{c})$ if $p(\overline{c}) \in I_w$;
for formulas $F,G$, $\tuple{I_H,I_T,w} \models F \wedge G$ if $\tuple{I_H,I_T,w} \models F$ and $\tuple{I_H,I_T,w} \models G$;
for formulas $F,G$, $\tuple{I_H,I_T,w} \models F \rightarrow G$ if $I_{w'} \models F \rightarrow G$ for all $w \leq w'$;
for a formula $F$, $\tuple{I_H,I_T} \models F$ if $\tuple{I_H,I_T,H} \models F$;
for a set of formulas $\Gamma$, $\tuple{I_H,I_T} \models \Gamma$ if $\tuple{I_H,I_T} \models F$ for all $F \in \Gamma$.
A HT-interpretation $\tuple{I_H,I_T}$ is a HT-model of a program $\Pi$ if $\tuple{I_H,I_T} \models \Gamma_\Pi$.
Let $\mathit{HT}(\Pi)$ be the set of HT-models of $\Pi$.
A HT-interpretation $\tuple{T,T}$ is an equilibrium model of $\Pi$ if $\tuple{T,T} \in \mathit{HT}(\Pi)$ and there is no $\tuple{H,T} \in \mathit{HT}(\Pi)$ with $H \subset T$ \cite{DBLP:journals/amai/Pearce06}.
Let $\mathit{EQ}(\Pi)$ be the set of equilibrium models of $\Pi$.

\begin{example}\label{ex:ht}
For $\Pi$ being the program in Example~\ref{ex:asp}, the theory $\Gamma_\Pi$ includes $e(1) \wedge (b(1) \rightarrow \bot) \rightarrow a(1)$ and other formulas.
For $X = \{e(1),e(2),a(1),b(2)\}$, $\mathit{HT}(\Gamma_\Pi)$ includes
$\tuple{X, X \cup \{\mathit{fail}\}}$, and no pair of the form $\tuple{I_H,X}$.
\hfill$\blacksquare$
\end{example}

\begin{proposition} % find reference, very likely Pearce 2004?
$I \in \mathit{SM}(\Pi)$ if and only if $\tuple{I,I} \in \mathit{EQ}(\Pi)$.
\end{proposition}

\section{Templates}\label{sec:templates}

A \emph{template} $\pi$ is a set of rules (like a program).
Let $\mathbf{P_L} \subseteq \mathbf{P}$ be the set of \emph{local predicates}, i.e., predicates whose name starts by double underscore.
Predicates in $\mathbf{P} \setminus \mathbf{P_L}$ are \emph{global} and play the role of (renamable) \emph{parameters} in templates.
% Local predicates starting by \lstinline|__static_| are also called \emph{static} 
% In the following, $\mathbf{P_L}$ denotes the set of predicates starting by double underscore;
%$, and $\mathbf{P_S}$ the set of predicates starting by \lstinline|__static_|;
% predicates in $\mathbf{P_L}$ are local predicates, 
%those in $\mathbf{P_S}$ are static (local) predicates, 
% and those in $\mathbf{P} \setminus \mathbf{P_L}$ are global predicates.
%
A \emph{renaming} $\rho$ is a function with signature $\rho : \mathbf{P} \longrightarrow \mathbf{P}$.
%, that is, $\rho$ replaces predicate names with possibly different predicate names.
A \emph{local renaming} $\rho$ is a renaming being the identity on predicates from $\mathbf{P} \setminus \mathbf{P_L}$.
%, that is, local renamings only apply to local predicates.
%Similarly, a \emph{static renaming} $\rho$ is a renaming being the identity on predicates from $\mathbf{P} \setminus \mathbf{P_S}$.
%, that is, static renamings only apply to static predicates.
In contrast, a \emph{global renaming} $\rho$ is a renaming being the identity on predicates from $\mathbf{P_L}$.
%, that is, global renamings do not apply to local predicates.
%
In the following, the term \emph{universally unique predicate} refers to a predicate name that is guaranteed to be unique, and the notation $\rho(\pi)$ is abused to refer to the set of rules in $\pi$ with all predicates renamed according to $\rho$;
similarly, $\rho(I)$ is the interpretation obtained from $I$ by renaming predicates according to $\rho$.
% \gibbi{For practical purposes, we will generate universally unique predicate names by appending pseudorandomly (?) generated UUIDs values~\cite{RFC4122} to local predicate names}. \comment{GB: dire qualcosa sulle proprietà degli UUID, es. l'irripetibilità virtuale}
% \malvi{Avevo pensato di tenere un po' astratto in questa sezione e poi nell'implementazione dire che usiamo gli UUID (cosa che poi ho dimenticato di dire). Adesso dovrebbe essere più chiaro (parte verde nell'implementazione). Ho anche messo UUID all'inizio di Background.}
% \fra{Con l'aggiunta della parte relativa agli UUID nella sezione background, credo che ora non sia più necessario aggiungere nulla qui.}

\begin{example}
The renaming $[\mathit{fail} \mapsto \mathit{\_\_fail}]$ is global;
it maps $\mathit{fail}$ to $\mathit{\_\_fail}$, and is the identity for other predicates.
The renaming $[\mathit{\_\_fail} \mapsto \mathit{\_\_fail\_7b905af5\_de82\_49b3\_9db7\_415d4d048c76}]$ is local;
with a very good probabilistic confidence the predicate $\mathit{\_\_fail\_7b905af5\_de82\_49b3\_9db7\_415d4d048c76}$ is unique if obtained appending a newly generated UUID to $\mathit{\_\_fail}$.
% (unless it is explicitly read after the UUID is generated).
% A static renaming could be $[\mathit{\_\_static\_fail} \mapsto \mathit{\_static\_fail\_4014fdf8\_554e\_4bc1\_85f0\_f3463923df2d}]$;
% note that $\mathit{\_static\_fail\_4014fdf8\ldots}$ is a global (yet unpredictable) name.
\hfill$\blacksquare$
\end{example}

%Each template $\pi$ is associated with a static renaming $\rho_\pi$ mapping every static predicate in $\mathit{pred}(\pi)$ to a universally unique global predicate in $\mathbf{P} \setminus \mathbf{P}_L$.
The \emph{application} $\pi\rho$ of a template $\pi$ w.r.t.\ a global renaming $\rho$ is the set of rules
$\rho(\rho_L(\pi))$,
%$\rho(\rho_L(\rho_\pi(\pi)))$,
where $\rho_L$ is a local renaming mapping each local predicate in $\mathit{pred}(\pi)$ to a universally unique predicate in $\mathbf{P_L}$.
%
%\comment{GB: in questo punto del testo il lettore non sa che un template consiste di regole e di template applications, e potrebbe non capire le considerazioni sul nesting. \malvi{la cosa è più semplice: l'applicazione di un template è un insieme di regole, quindi una volta che lo espandi lo puoi avere come parte di un altro template (che è un insieme di regole). Ho riformulato.}}
Note that, by the way they are defined, a template $\pi$ (being a set of rules) can include the application of another template $\pi'\rho$ (i.e., a set of rules):
in this case, $\pi \supseteq \pi'\rho$, and 
% $\mathit{pred}(\pi'\rho) \cap \mathbf{P_L} \cap \mathit{pred}(\pi \setminus \pi'\rho) = \emptyset$
the sets of local predicates occurring in $\pi'\rho$ and $\pi \setminus \pi'\rho$ are disjoint because those in $\pi'\rho$ are universally unique by construction;
moreover, any application $\pi\rho'$ of $\pi$, by construction, is guaranteed to map local predicates of $\pi'\rho$ to new universally unique predicates, hence preserving the invariant that different applications of $\pi$ are associated with different local predicates.
% Note that templates can be non-recursively nested because the application of a template is a set of rules, and local predicates of deeper templates are mapped to local predicates, so that multiple applications of the highest template still use different names for such predicates.
%in contrast, note that static local predicates are directly mapped to global predicate names so that they are shared among all instances of a template, even if it is instantiated deep inside another template.
% Finally, in the following the term program is used to refer to a template application not being part of other templates.

\begin{example}\label{ex:template-1}
Let $\pi_\mathit{tc}$ (for transitive closure) be the template comprising of
\begin{align}
%    \mathit{c}(X,Y) & {}\leftarrow \mathit{r}(X,Y) \label{eq:ex:template-1:1}\\
%    \mathit{c}(X,Z) & {}\leftarrow \mathit{c}(X,Y),\ \mathit{r}(Y,Z) \label{eq:ex:template-1:2}
    \mathit{c}(X,Y) \leftarrow \mathit{r}(X,Y) \qquad & \qquad \mathit{c}(X,Z) \leftarrow \mathit{c}(X,Y),\ \mathit{r}(Y,Z) \label{eq:ex:template-1:1}%\label{eq:ex:template-1:2}
\end{align}
that is, $\pi_\mathit{tc}$ is expected to define the transitive closure of the binary relation encoded by predicate $r$.
The application
$\pi_\mathit{tc}[\mathit{r} \mapsto \mathit{link}, \mathit{c} \mapsto \mathit{reach}]$
is expected to produce (at least) the transitive closure of $\mathit{link/2}$ in predicate $\mathit{reach/2}$.
Similarly, the application
$\pi_\mathit{tc}[\mathit{r} \mapsto \mathit{link}, \mathit{c} \mapsto \mathit{link}]$
is expected to enforce that relation $\mathit{link/2}$ is closed under transitivity.
Let $\pi_\mathit{tcc}$ (for transitive closure check) comprise of the
rule $\bot \leftarrow \mathit{\_\_c}(X,X)$
and the application
$\pi_\mathit{tc}[\mathit{c} \mapsto \mathit{\_\_c}]$,  i.e., the rules
\begin{align}
    % \mathit{\_\_c}(X,Y) & {}\leftarrow \mathit{r}(X,Y) \label{eq:ex:template-2:1}\\
    % \mathit{\_\_c}(X,Z) & {}\leftarrow \mathit{\_\_c}(X,Y),\ \mathit{r}(Y,Z) \label{eq:ex:template-2:2}\\
    \mathit{\_\_c}(X,Y) \leftarrow \mathit{r}(X,Y) \qquad & \qquad
    \mathit{\_\_c}(X,Z) \leftarrow \mathit{\_\_c}(X,Y),\ \mathit{r}(Y,Z) \label{eq:ex:template-2:2}
    % \bot & {}\leftarrow \mathit{\_\_c}(X,X) \label{eq:ex:template-2:3}
\end{align}
The template $\pi_\mathit{tcc}$ is expected to discard interpretations in which the relation encoded by $\mathit{r/2}$ has cycles.
Predicate $\mathit{c}$ of template $\pi_\mathit{tc}$ is mapped to a local predicate of $\pi_\mathit{tcc}$ so to inhibit external reference.
The template application $\pi_\mathit{tcc}[\mathit{r} \mapsto \mathit{link}]$ could map
$\mathit{\_\_c}$ to $\mathit{\_\_c\_6bd3728a\_36b4\_4fb9\_8019\_61af6363420b}$.
% \begin{align*}
%     \mathit{\_\_c\_6bd3728a...}(X,Y) & {}\leftarrow \mathit{link}(X,Y) \\
%     \mathit{\_\_c\_6bd3728a...}(X,Z) & {}\leftarrow \mathit{\_\_c\_6bd3728a...}(X,Y),\ \mathit{r}(Y,Z) \\
%     \bot & {}\leftarrow \mathit{\_\_c\_6bd3728a...}(X,X)
% \end{align*}
% where $\mathit{\_\_c\_6bd3728a...}$ is actually $\mathit{\_\_c\_6bd3728a\_36b4\_4fb9\_8019\_61af6363420b}$.
Let $\pi_\mathit{tcg}$ (for transitive closure guaranteed) comprise $\pi_\mathit{tc}[\,] \cup \pi_\mathit{tc}[\mathit{c} \mapsto \mathit{\_\_c}]$ and
$\bot \leftarrow \mathit{c}(X,Y),\ \naf \mathit{\_\_c}(X,Y)$,
that is, the latter rule and \eqref{eq:ex:template-1:1}--\eqref{eq:ex:template-2:2}.
Template $\pi_\mathit{tcg}$ computes the transitive closure of $\mathit{r/2}$ in $\mathit{c/2}$, and enforces failure if $\mathit{c}$ is extended externally with possible rule additions mentioning $\mathit{c}$.
\hfill$\blacksquare$
\end{example}

\section{Properties}\label{sec:properties}

Some properties of templates based on their HT-models can be used to establish invariants on the stable models of broader programs.
We first consider a pair $\Pi,\Pi'$ of programs, their HT-models, and possibly the fact that some predicates occur only in $\Pi$.
Later on, we recast the results for templates.
Let us start with simple conditions guaranteeing that some atoms have fixed truth values in all stable models of programs extending $\Pi$.
Intuitively, as the possible interpretations of world $T$ provide an overestimate on stable models, their intersection and union can be analyzed to identify atoms that are necessarily true or false in all (stable) models;
%\comment{GB - question: possono esserci atomi falsi in tutti gli AS che però compaiono nell'unione dei modelli T? Congetturo di sì. In tal caso possiamo identificare solo una parte dei sempre falsi (ad intuito questi falsi che possiamo identificare sono i falsi del modello well-founded, ma non sono sicuro). Dobbiamo dire qualcosa.
%\malvi{Sì, Proposition~3 ti dà alcuni atomi che sono sicuramente falsi anche se ci sono modelli in cui sono veri; ci possono essere altri atomi sicuramente falsi che non riesci a riconoscere con Proposition~3 (direi già se forzi la falsità di questi atomi puoi trovare altri veri/falsi rispetto ai modelli e ripetere). Non credo che se un atomo è falso nel modello well-founded di $\Pi$ lo puoi mettere a falso negli AS di $\Pi \cup \Pi'$ (WF è non-monotono); ad esempio, se hai \lstinline|a :- b| il modello WF è falso, ma se aggiungi il fatto b allora a e b sono veri.}
%};
see \eqref{eq:true-inv}--\eqref{eq:false-inv} below.
On the other hand, for a fixed interpretation of world $T$, the possible interpretations of world $H$ provide an overestimate on the models of a program reduct, and therefore their intersection can be analyzed to identify atoms that are necessarily true in the reduct;
see \eqref{eq:reduct-inv} below.

% garanzia su assegnamento atomi in tutti gli answer set
\begin{proposition}\label{prop:models}
Let $\Pi,\Pi'$ be programs.
It holds that 
\begin{align}
    \bigcap_{\tuple{I_H,I_T} \in \mathit{HT}(\Gamma_\Pi)}{I_T} & {}\subseteq \bigcap_{\tuple{I_H,I_T} \in \mathit{HT}(\Gamma_\Pi \cup \Gamma_{\Pi'})}{I_T} \label{eq:true-inv} \\
    \bigcup_{\tuple{I_H,I_T} \in \mathit{HT}(\Gamma_\Pi)}{I_T} & {}\supseteq \bigcup_{\tuple{I_H,I_T} \in \mathit{HT}(\Gamma_\Pi \cup \Gamma_{\Pi'})}{I_T}. \label{eq:false-inv}
\end{align}
Moreover, for any interpretation $I_T$, it holds that 
\begin{align}
    \bigcap_{\tuple{I_H,I_T} \in \mathit{HT}(\Gamma_\Pi)}{I_H} & {}\subseteq \bigcap_{\tuple{I_H,I_T} \in \mathit{HT}(\Gamma_\Pi \cup \Gamma_{\Pi'})}{I_H}. \label{eq:reduct-inv}
    % l'unione non ha senso perché è sempre $I_T$
    % \bigcup_{\tuple{I_H,I_T} \in \mathit{HT}(\Gamma_\Pi)}{I_H} & {}\supseteq \bigcup_{\tuple{I_H,I_T} \in \mathit{HT}(\Gamma_\Pi \cup \Gamma_{\Pi'})}{I_H} \label{eq:false-reduct-inv}
\end{align}
\end{proposition}

From \eqref{eq:true-inv}--\eqref{eq:false-inv} we obtain \eqref{eq:assignment-inv} below,
and from \eqref{eq:reduct-inv} we obtain \eqref{eq:reduct-true-inv} below.

\begin{corollary}\label{cor:models}
Let $\Pi,\Pi'$ be programs, and $I,J$ be interpretations with $J \subseteq I$.
\begin{align}
    I \models \Pi \cup \Pi' \Longrightarrow \bigcap_{\tuple{I_H,I_T} \in \mathit{HT}(\Gamma_\Pi)}{I_T} \subseteq I \subseteq \bigcup_{\tuple{I_H,I_T} \in \mathit{HT}(\Gamma_\Pi)}{I_T}. \label{eq:assignment-inv} \\
    I \models \Pi \cup \Pi' \wedge J \models (\Pi \cup \Pi')^I \Longrightarrow \bigcap_{\tuple{I_H,I_T} \in \mathit{HT}(\Gamma_\Pi)}{I_H} \subseteq J. \label{eq:reduct-true-inv}
\end{align}
\end{corollary}

\begin{example}
Let $\Pi$ be the following set of rules:
\begin{lstlisting}
$\derives$ a(X).  b(1).  g $\derives$ b(X), not a(X).  $\derives$ not d.  e $\derives$ not f. f $\derives$ not e.
\end{lstlisting}
For any $\Pi'$, atom \lstinline|a($c$)| is false in all models of $\Pi \cup \Pi'$, for all $c \in \mathbf{C}$;
indeed, one can see that \lstinline|a($c$)| is false in all $I_T$ such that $\tuple{I_H,I_T} \in \mathit{HT}(\Gamma_\Pi)$, and therefore \eqref{eq:assignment-inv} applies.
Similarly, \lstinline|b(1)| is true in all models of $\Pi \cup \Pi'$ and their reducts;
indeed, \lstinline|b(1)| is true in all $I_H$ and $I_T$ such that $\tuple{I_H,I_T} \in \mathit{HT}(\Gamma_\Pi)$, and therefore \eqref{eq:assignment-inv}--\eqref{eq:reduct-true-inv} apply.
We can go on and conclude that \lstinline|g| is true in all models of $\Pi \cup \Pi'$ and their reducts,
and that \lstinline|d| is true in all models of $\Pi \cup \Pi'$ (but not necessarily in their reducts).
Finally, it can be checked that \lstinline|e| is true in all models of $(\Pi \cup \Pi')^I$ such that $I \models \Pi \cup \Pi'$ and \lstinline|e| belongs to $I$ (similar for \lstinline|f|);
\eqref{eq:reduct-true-inv} applies.
\hfill$\blacksquare$
\end{example}

As shown in the next proposition, further interpretations can be guaranteed to not be stable models of $\Pi \cup \Pi'$:
the truth value of atoms whose predicates are guaranteed to occur in $\Pi$ only, cannot compromise the satisfiability of $\Pi'$.
Hence, if the instability of a model of $\Pi$ only depends on such predicates, the instability extends to $\Pi \cup \Pi'$.

% questo ci dà impossibilità di answer set (ad esempio, si può simulare un constraint con __fail :- not __fail, foo)
\begin{proposition}\label{prop:unstable}
Let $\Pi,\Pi'$ be programs, and $X$ be a nonempty set of atoms of the form $p(\overline{c})$, with $p \in \mathit{pred}(\Pi) \setminus \mathit{pred}(\Pi')$.
If $\tuple{I,I \cup X} \in \mathit{HT}(\Gamma_{\Pi})$, then $I \cup X \notin \mathit{SM}(\Pi \cup \Pi')$.
\end{proposition}

\begin{example}\label{ex:unstable}
Let $\Pi$ be
\lstinline|__fail $\derives$ foo, not __fail.|
Note that $\mathit{HT}(\Pi)$ includes 
$\tuple{\{\text{\lstinline|foo|}\} \cup X, \{\text{\lstinline|foo, __fail|}\} \cup X}$
and
$\tuple{X, \{\text{\lstinline|__fail|} \cup X\}}$,
for every set $X$ of atoms not including \lstinline|foo| or \lstinline|__fail|.
For every $\Pi'$ not mentioning \lstinline|__fail|, $\mathit{SM}(\Pi \cup \Pi')$ cannot contain $\{\text{\lstinline|__fail|}\} \cup X$ and $\{\text{\lstinline|foo, __fail|}\} \cup X$;
Proposition~\ref{prop:unstable} applies.
\hfill$\blacksquare$
\end{example}

Elaborating on the above claim, it is possible to conclude that a specific set $I$ of atoms cannot be extended to a stable model without including at least one atom in another given set $I'$.
As a special case, when $I = \{\alpha\}$ and $I' = \emptyset$, falsity of $\alpha$ is guaranteed in all stable models;
this is the case for \lstinline|__fail| in Example~\ref{ex:unstable}.

\begin{corollary}\label{cor:unstable}
Let $\Pi,\Pi'$ be programs, and be $I,I'$ disjoint sets of atoms.
If every $I_T$ with $I \subseteq I_T$ and $I' \cap I_T = \emptyset$ is such that there is 
$\tuple{I_H,I_T} \in \mathit{HT}(\Gamma_{\Pi})$ with $I_H \subset I_T$ and $p \in \mathit{pred}(\Pi) \setminus \mathit{pred}(\Pi')$ for all $p(\overline{c}) \in I_T \setminus I_H$, then there is no $I_T \in \mathit{SM}(\Pi \cup \Pi')$ with $I \subseteq I_T$ and $I' \cap I_T = \emptyset$.
% Let $\Pi,\Pi'$ be programs, and $\alpha$ be a ground atom.
% If every $\tuple{I_H,I_T} \in \mathit{HT}(\Gamma_{\Pi})$ with $\alpha \in I_T$ is such that $I_H \subset I_T$ and $p \in \mathit{pred}(\Pi) \setminus \mathit{pred}(\Pi')$ for all $p(\overline{c}) \in I_T \setminus I_H$, then there is no $I \in \mathit{SM}(\Pi \cup \Pi')$ with $\alpha \in I$.
\end{corollary}

All in all, given a program $\Pi$ whose local predicate names are guaranteed to be universally unique, we are interested in the following test types on $\Pi$:
\begin{enumerate}
\item[\bf T1.]
Given sets $I,J$ of atoms, apply Corollary~\ref{cor:models} to verify that atoms in $I$ are true in all (classical) models of $\Pi \cup \Pi'$, and atoms in $J$ are false in all models of $\Pi \cup \Pi'$, where $\Pi'$ is any program.

\item[\bf T2.]
Given a model $I$ of $\Pi$, and a set $J \subseteq I$, apply Corollary~\ref{cor:models} to verify that atoms in $J$ are true in all models of $(\Pi \cup \Pi')^I$, where $\Pi'$ is any program.

\item[\bf T3.]
Given disjoint sets $I,I'$ of atoms, apply Corollary~\ref{cor:unstable} to verify that there is no $I \cup X \in \mathit{SM}(\Pi \cup \Pi')$ such that $X \cap I' \neq \emptyset$, for any set $X$ of atoms and program $\Pi'$
(i.e., some atom in $I'$ must be true when atoms in $I$ are true).
\end{enumerate}
Note that the above tests are sound, and not intended to be complete. For instance, it is possible that there is no $I \cup X \in \mathit{SM}(\Pi \cup \Pi')$ such that $X \cap I' \neq \emptyset$, for any set $X$ of atoms and program $\Pi'$, but this is not captured by Corollary~\ref{cor:unstable}.
Finally, template instantiation guarantees that local predicate names are universally unique.

% \comment{GB: siamo sicuri di volerci fare un teorema? Entrano in gioco le proprietà degli UUID, che sono ottime per tutti i fini pratici, ma c'è sempre la probabilità non nulla di una collisione. Oppure formuliamo il teorema "under assumption of uniqueness of UUIDs"}
\begin{theorem}\label{thm:local}
Let $\pi$ be a template, %such that all head occurrences of predicates in $\mathit{pred}(\pi) \cap \mathbf{P_S}$ are facts.
$\rho$ be a global renaming $\rho$, and $\Pi = \pi\rho$.
The requirement $p \in \mathit{pred}(\Pi) \setminus \mathit{pred}(\Pi')$ in Proposition~\ref{prop:unstable} and Corollary~\ref{cor:unstable} is guaranteed for predicates in $\mathit{pred}(\pi\rho) \cap \mathbf{P_L}$,
under the assumption that generated UUIDs are unique and $\Pi'$ has zero knowledge of the local renaming used by $\pi\rho$.
\end{theorem}

\section{Implementation}\label{sec:impl}

Templates are implemented in \lstinline|dumbo-asp| (\url{https://github.com/alviano/dumbo-asp}), a prototype \textsc{python} library powered by the \textsc{clingo python api} \cite{DBLP:journals/tplp/KaminskiRSW23};
\lstinline|dumbo-asp| is mainly intended to be used as an API, particularly regarding the definition of automated tests that are expected to be written in acknowledged frameworks like \textsc{pytest}.
Nonetheless, in order to smooth out the learning curve for developers more acquainted with ASP than other programming languages, the tool supports also a serialization format based on ASP rules to declare and apply templates. 

Predicates \lstinline|__apply_template__|, \lstinline|__template__| and \lstinline|__end__| are reserved.
A \emph{program with templates} is a sequence $[\pi_1,\ldots,\pi_n]$ ($n \geq 0$), where each $\pi_i$ is one of the followings:
% \begin{itemize}
% \item 
(i) a rule of the form \eqref{eq:rule};
%
% \item 
(ii) a template application, that is, % a fact of the form 
\begin{lstlisting}
    __apply_template__("$\mathit{name}$", $\mathit{mapping}$).
\end{lstlisting} 
where $\mathit{name}$ identifies a template occurring at some previous index $j < i$, and $\mathit{mapping}$ is given by a comma-separated list of pairs of the form $(\mathit{old}, \mathit{new})$;
%
% \item
(iii) a template declaration, that is, a block % of the form
\begin{lstlisting}
    __template__("$\mathit{name}$").   $\mathit{content}$   __end__.
\end{lstlisting} 
where $\mathit{name}$ identifies the template,
%(we suggest to adopt a schema like \lstinline|@user/name| to ease cooperation among different developers), 
and $\mathit{content}$ is a sequence of rules % of the form \eqref{eq:rule} 
and applications of previous templates. % occurring at index smaller than $i$.
% \end{itemize}
Note that recursive template applications are disallowed by design, but arbitrary dependencies among predicates defined by different template applications are permitted, including recursion. %, as it will be clarified by the expansion procedure.

\begin{algorithm}[t]
    \caption{Expand($[\pi_1,\ldots,\pi_n]$: a program with templates; $\phantom{xxxx}$ $\mathit{templates}$: a map from names to templates): a program $\Pi$}\label{alg:expand}
    $\Pi := \emptyset$\;
    \ForEach{$i \in 1..n$}{
        \uIf{$\pi_i$ is \lstinline|__template__("$\mathit{name}$").| $\mathit{content}$ \lstinline|__end__.|}{
            $\mathit{templates}[\mathit{name}] := \mathrm{Expand}(\mathit{content}, \mathit{templates})$\;
        }
        \uElseIf{$\pi_i$ is \lstinline|__apply_template__("$\mathit{name}$", $\rho$).|}{
            $\pi := \mathit{templates}[\mathit{name}]$;\qquad
            $\Pi := \Pi \cup \pi\rho$\;
        }
        \Else{
            $\Pi := \Pi \cup \{\pi_i\}$\;
        }
    }
    \Return{$\Pi$}\;
\end{algorithm}
A program with templates can be expanded by Algorithm~\ref{alg:expand}.
Elements of the program are processed in order (line~2).
Ordinary rules are added to the output program (line~8), which is initially empty (line~1).
If a template declaration is found (lines~3--4), the $\mathit{templates}$ map (initially containing built-in templates from our core library) is extended with a template comprising rules obtained by calling Algorithm~\ref{alg:expand};
note that the nested call to the algorithm is not reiterated thanks to the serialization format given above (templates' content cannot declare other templates).
Whenever the application of a template is found, the content of the template is retrieved from the $\mathit{templates}$ map, and the global renaming $\rho$ is used to produce rules %$\rho(\rho_L(\rho_\pi(\pi)))$
$\rho(\rho_L(\pi))$
for the output program (lines~3--4);
in our implementation, %$\rho_\pi(\_\_\mathit{static}\_p) = \_\mathit{static}\_p\_u$, where $u$ is a UUID generated when $\pi$ is declared, and 
$\rho_L(\_\_p) = \_\_p\_u$, where $u$ is a UUID generated when $\pi$ is applied.

\begin{example}[Continuing Example~\ref{ex:template-1}]\label{ex:program-with-templates}
Let us consider the following program:
\begin{lstlisting}[numbers=left]
__template__("@d/tc").
    c(X,Y) $\derives$ r(X,Y).    c(X,Z) $\derives$ c(X,Y), r(Y,Z).
__end__.

__template__("@d/tcg").
    __apply_template__("@d/tc", (c, __c)).
    c(X,Y) $\derives$ __c(X,Y).     $\derives$ c(X,Y), not __c(X,Y).
__end__.

link(a,b). link(a,c). __apply_template__("@d/tcg", (r, link), (c,reach)).
reach(foo,bar).  % this is going to cause an inconsistency
\end{lstlisting}
Template \lstinline|@d/tcg| materializes the transitive closure in the local predicate \lstinline|__c| by applying \lstinline|@d/tc| (line~5); \lstinline|__c| is then \dquo{copied} to the global predicate \lstinline|c|, subject to a constraint guaranteeing that it cannot be further extended elsewhere (line~6).
In fact, line~9 causes an inconsistency with such an invariant of the program.
Also note that these templates are part of the core templates of \lstinline|dumbo-asp|, which however use longer, more understandable names (\lstinline|relation| instead of \lstinline|r|, \linebreak \lstinline|@dumbo/transitive closure| instead of \lstinline|@d/tc|, and so on).
\hfill$\blacksquare$
\end{example}

Regarding test types defined in Section~\ref{sec:properties}, we expect $\Pi$ to be the application of a template possibly extended with other rules providing a context for specific behaviors of the template.
Our implementation is powered by meta encodings coupled with the \emph{reification} of $\Pi$ \cite{DBLP:journals/tplp/KaminskiRSW23}, as well as other rules to check for specific conditions.
Tests of type \textbf{T1} are implemented by computing the intersection and union of all models of $\Pi$ by means of cautious and brave reasoning.
Type \textbf{T2} is implemented by computing the intersection of all models of $\Pi^I$ by means of cautious reasoning.
\textbf{T3} tests are implemented by enumerating the models of $\Pi$ including $I$ and disjoint from $I'$, and for each of them checking that there is a model for their reduct satisfying the requirements of Corollary~\ref{cor:unstable};
both computational tasks are addressed by stable model search.
We provide functions raising exceptions if some of $\mathbf{T1}$--$\mathbf{T3}$ fail:
\begin{lstlisting}
validate_in_all_models(program, true_atoms, false_atoms)
validate_in_all_models_of_the_reduct(program, model, true_atoms)
validate_cannot_be_extended_to_stable_model(prg, true_atoms, false_atoms)
\end{lstlisting}
% Testing frameworks like \textsc{pytest} can easily check for raising of exceptions, as well as for the non-raising of expected exceptions.
% Note that \emph{unknown atoms} can be specified to prevent grounding simplifications for atoms that are relevant for the test, but whose truth value is not.
% \comment{GB: questa ultima frase sottointende dettagli implementativi sul grounder sottostante che non abbiamo dato, la potremmo togliere.}

\section{Application Scenario}\label{sec:use-case}

\begin{figure}[t]
    \centering
    \includegraphics[width=.6\textwidth]{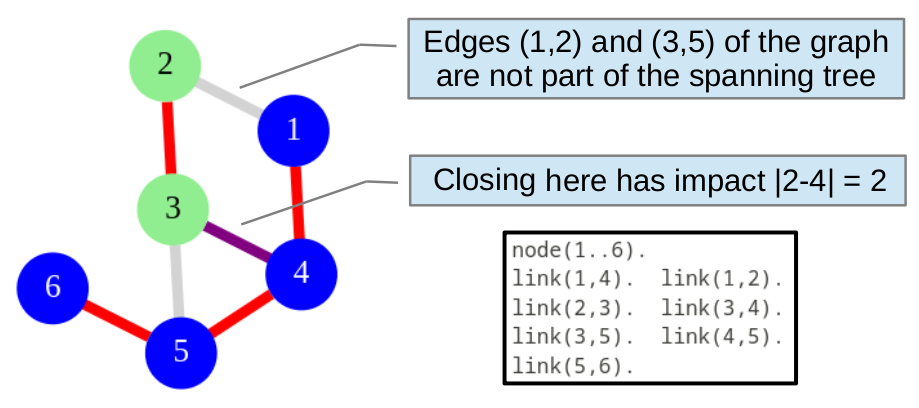}
    \caption{Example of input and output for the running problem of Section~\ref{sec:use-case}}
    \label{fig:input-output}
\end{figure}

Let us consider a hypothetical (partial) problem specification to be addressed by two teams of developers, say \emph{Alpha} and \emph{Bravo}.
Given a graph representing road segments, we are interested in finding a spanning tree to build a highway network.
For each such network proposal, we want to understand the impact of closing every single road segment in terms of the resulting tree-size-difference between connected points. An example input graph, one of its spanning trees and the impact of closing one of its segments are shown in Figure~\ref{fig:input-output}.

\begin{figure}[t]
    \begin{minipage}{.45\textwidth}
\begin{lstlisting}[numbers=left, basicstyle=\tt\scriptsize]
link(X,Y) $\derives$ link(Y,X).

{tree(X,Y) : link(X,Y), X < Y} = C-1           $\derives$ C = #count {X : node(X)}.
tree(X,Y) $\derives$ tree(Y,X).
reach(X) $\derives$ X = #min {Y : node(Y)}.
reach(Y) $\derives$ reach(X), tree(X,Y).
$\derives$ node(X), not reach(X).
\end{lstlisting}
    \end{minipage}
    \begin{minipage}{.50\textwidth}
\begin{lstlisting}[numbers=left, basicstyle=\tt\scriptsize]
{out(X,Y) : tree(X,Y)} = 1.
in(X,Y) $\derives$ tree(X,Y), not out(X,Y).
in(X,Y) $\derives$ in(Y,X).
reach(X) $\derives$ X = #min {Y : node(Y)}.
reach(Y) $\derives$ reach(X), in(X,Y).
impact(X,Y,|C|) $\derives$ out(X,Y), C = #sum{1,Z : reach(Z); -1,Z : node(Z), not reach(Z)}.
\end{lstlisting}        
    \end{minipage}
    
    \caption{Programs written by Team Alpha (left) and Team Bravo (right)}\label{fig:naive}
\end{figure}

Team Alpha develops a declarative model for spanning trees, and Team Bravo develops the impact measurement.
The two teams agree on using predicates \lstinline|node/1| and \lstinline|link/2| for the input graph, \lstinline|tree/2| for the spanning tree, and \lstinline|impact/3| for measuring the impact of closing one segment.
The two teams produce respectively the ASP-Core-2 programs in Figure~\ref{fig:naive}.
Taken individually, the two programs are correct, which is not the case for their union because \lstinline|reach/1| is used with different meanings;
after some synchronization between the two teams, the bug is fixed by changing \lstinline|reach/1| to \lstinline|reach'/1| in one of the two programs.
Besides this, there is another bug due to the fact that Alpha enforces the symmetric closure of \lstinline|tree/2|, while Bravo works under the assumption that \lstinline|tree/2| is anti-symmetric;
the bug can be fixed by adding \lstinline|X < Y| in lines~1--2 of Bravo.
Moving the code to a new project may lead to similar issues, especially for very common predicate names like \lstinline|in/2| and \lstinline|out/2|.
In addition, observe that some rules are essentially repeating (e.g., lines~4--5 of Alpha and lines~4--5 of Bravo).

\begin{figure}[t]
\begin{lstlisting}[numbers=left, basicstyle=\tt\scriptsize]
__template__("@d/symmetric closure").
    c(X,Y) $\derives$ r(Y,X).    c(X,Y) $\derives$ r(Y,X).
__end__.

__template__("@d/reachable nodes").
    reach(X) $\derives$ start(X).    reach(Y) $\derives$ reach(X), link(X,Y).
__end__.

__template__("@d/connected graph").
    __start(X) $\derives$ X = #min{Y : node(Y)}.
    __apply_template__("@d/reachable nodes", (start, __start), (reach, __reach)).
    $\derives$ node(X), not __reach(X).
__end__.


__apply_template__("@d/symmetric closure", (r, link), (c, link)).

__template__("spanning tree").
    {tree(X,Y) : link(X,Y), X < Y} = C-1 $\derives$ C = #count{X : node(X)}.
    __apply_template__("@d/symmetric closure", (r, tree), (c, __tree)).
    __apply_template__("connected graph", (node, node), (link, __tree)).
__end__.

__apply_template__("spanning tree").
\end{lstlisting}
    
    \caption{Program written by Team Alpha (lines~13--19) using core templates (lines~1--12)}\label{fig:alpha}
\end{figure}

\begin{figure}[t]
\begin{lstlisting}[numbers=left, basicstyle=\tt\scriptsize]
__apply_template__("@d/symmetric closure", (r, tree), (c, tree)).

{__out(X,Y) : tree(X,Y)} = 1.
__in(X,Y) $\derives$ tree(X,Y), not __out(X,Y).
__apply_template__("@d/symmetric closure", (r, __in), (c, __in)).

__start(X) $\derives$ X = #min{Y : node(Y)}.
__apply_template__("@d/reachable nodes", (start,__start), (link, __in), (reach,__reach)).
impact(X,Y,|C|) $\derives$ __out(X,Y), C = #sum{1,Z : __reach(Z); -1,Z : node(Z), not __reach(Z)}.
\end{lstlisting}        
    
    \caption{Program written by Team Bravo using core templates}\label{fig:bravo}
\end{figure}

Let us consider a different development timeline.
Alpha is aware of templates and the program in Figure~\ref{fig:alpha} is produced (where relevant core templates from our library are also shown for convenience). Bravo is not aware of templates and follows the traditional ASP development lifecycle, using a plain solver of choice.
%The template \lstinline|connected graph| written by Alpha is also associated with several tests to verify that reachable nodes are properly computed, graph connectivity is properly checked, and spanning trees are correctly computed (tests not shown here).
%\gibbi{ qui sicuro ci chiedono 'how the test would look like', e con solo set di atomi in input ai test non possiamo davvero modellare questi invariant. O mi sono perso qualche passaggio?}
%\malvi{sì, intendo per alcuni testcase puoi controllare che il comportamento è quello atteso e volendo si possono estendere un po' i template per controllare che non c'è interferenza esterna, ma meglio non dirlo (ho tolto la frase)}
Alpha is ready to share their code with Bravo, actually in the form of an expanded, transpiled program obtained by Algorithm~\ref{alg:expand}, so that a clash of names is essentially impossible. In this timeline Bravo can use the transpiled code of Alpha without installing any additional software.
This timeline may evolve with Bravo liking the idea of templates, and reusing some of the templates written by Alpha.
The result is shown in Figure~\ref{fig:bravo}.
The two teams may also add closure constraints to guarantee that the extension of \lstinline|tree/2| and \lstinline|impact/3| is not accidentally extended by other external rules;
for this purpose, we provide templates \lstinline|@d/exact copy (arity $n$)| for $n \geq 0$, which have the following form:
\begin{lstlisting}
__template__("@d/exact copy (arity $n$)").
    output(X1,...,X$n$) :- input(X1,...,X$n$).
    :- output(X1,...,X$n$), not input(X1,...,X$n$).
__end__.
\end{lstlisting}
Even better, as \lstinline|reachable nodes|, \lstinline|connected graph| and \lstinline|spanning tree| are very likely reusable in other programs, Alpha may propose their addition to the core library, or publish them elsewhere.

\section{Related Work}\label{sec:rw}

Our work has clear points of connection with three, not necessarily disjoint, lines of research: {\em a)} studies on modular ASP, {\em b)} practical approaches at verifying, debugging and unit testing ASP programs, and {\em c)} studies on relativized equivalence of logic programs under stable models semantics.

Regarding \emph{a)}, {\em modular extensions to ASP} are historically classified in 
\emph{programming-in-the-large} approaches, where the focus is on the composition of arbitrary sets of rules~\cite{DBLP:journals/jair/JanhunenOTW09}, with no explicit notion of scope, and \emph{programming-in-the-small} approaches, where some form of scoping and notions of input/output predicates are proposed.
The proposal of generalized quantifiers in~\cite{DBLP:conf/lpnmr/EiterGV97}, macros in~\cite{DBLP:conf/iclp/BaralDT06}, templates in~\cite{DBLP:journals/aicom/CalimeriI06} and module atoms in~\cite{DBLP:conf/iclp/Dao-TranEFK09} fall in this latter category,
while multi-context systems~\cite{DBLP:conf/kr/Dao-TranEFK10} feature aspects of both approaches.
It must be noted that we propose a mixed approach which is mainly based on macro expansion, yet bringing aspects of programming-in-the-large.
In particular, within a template we do not require an explicit distinction between input and output predicates, and definitions of predicates are not confined to the template. 
This is in contrast with macros and the previous proposal of templates, where input and output relations need to be specified ahead;
moreover, in previous works name clashes were not explicitly addressed, although it was hinted at weaker handling of this issue without providing an actual invariant in this respect, especially in case transpiled code is moved in other projects.
Note also that we aim to reuse templates in combination with future, unknown, logic specifications: 
in a way, we aim to compose programs from smaller bricks, in a bottom-up fashion. 
Among the modular approaches, a somewhat orthogonal, top-down methodology has been proposed by Cabalar et al.~\cite{DBLP:journals/tplp/CabalarFL20}, where it is suggested that single logic programs, built by individual knowledge designers, can be devised in a modular structure.
The correctness of such program parts, expressed in terms of a form of strong equivalence, helps in verifying the entire module structure (i.e. the original program).

% end NOP

Concerning \emph{b)}, practical approaches to debugging and testing in the context of ASP, such as unit tests and TDD, have been considered mainly at the level of easing the embedding of test cases within a program. In this respect, linguistic extensions have been proposed to specify that some rules extended with a provided set of facts are expected to produce a stable model, or on the contrary that some stable models are not expected \cite{DBLP:conf/jelia/AmendolaBR21,DBLP:conf/cilc/FebbraroRR11,DBLP:conf/lpnmr/GresslerOT17,DBLP:conf/ecai/JanhunenNOPT10,DBLP:conf/lpnmr/JanhunenNOPT11,DBLP:conf/kr/OetschPPST12,DBLP:journals/tplp/VosKOPT12}.
While it is clear that such linguistic extensions provide valuable tools for developers, they are not meant to guarantee that a set of rules can be used in another program still behaving in a controlled way. In part, this is due to the nonmonotonicity of stable model semantics, but there are also assumptions that cannot be enforced, among them the fact that a predicate is not used elsewhere.
Another way of checking properties of a program is by defining \emph{achievements}~\cite{DBLP:journals/tplp/Lifschitz17}, that is, statements on the behavior of the first $n$ rules of a program (said \emph{prefixes}), for some $n \geq 1$.
While achievements can be given in terms of first-order logic assertions, and can be automatically verified for linguistic fragments of ASP, by design they cannot be used to check properties of any portion of a program not being a prefix.
Actually, the properties of a prefix of the program may be lost when other rules of the program are added, possibly due to the very last considered rules.
Active research in this context led to the release of the \textsc{anthem} tool \cite{DBLP:journals/tplp/FandinnoLLS20}, enabling the possibility of verifying that \emph{io-programs} conform to first-order specifications, where an io-program is essentially a program with distinguished input and output predicates;
input predicates only occur in rule bodies, and predicates not being input or output are called private.
Since our templates provide a simple mechanism to guarantee that local predicates are essentially private, \textsc{anthem} can be employed to verify some of their properties.
The idea is to not use input predicates in rule heads, define all relations using local predicates, and finally define output predicates by applying the \lstinline|@d/exact copy (arity $n$)| templates.

\begin{example}
Recall the \lstinline|spanning tree| template shown in Figure~\ref{fig:alpha}.
Let $\Pi$ be
\begin{lstlisting}
__apply_template__("spanning tree", (tree, __t)).
__apply_template__("@d/exact copy (arity 2)",(input,__t),(output,tree)).
\end{lstlisting}
The application of $\Pi$ w.r.t.\ the identity renaming, $\Pi[\,]$, is
\begin{lstlisting}
{__t(X,Y) : link(X,Y), X < Y} = C-1 :- C = #count{X : node(X)}.

__t_ef9...(X,Y) :- __t(X,Y).  % symmetric closure
__t_ef9...(X,Y) :- __t(Y,X).  % symmetric closure

% connectedness
__start_a48..._ef9...(X) :- X = #min{Y : node(Y)}.
__reach_a48..._ef9...(X) :- __start_a48..._ef9...(X).
__reach_a48..._ef9...(Y) :- __reach_a48..._ef9...(X), __t_ef9...(X,Y).
:- node(X), not __reach_a48473e1..._ef9...(X).

tree(X0,X1) :- __t(X0,X1).   :- tree(X0,X1); not __t(X0,X1).  % output
\end{lstlisting}
It can be noted that $\Pi[\,]$ is essentially an io-program with input predicates \lstinline|node/1| and \lstinline|link/2|, and output predicate \lstinline|tree/2|.
Other predicates are private.
\hfill$\blacksquare$
\end{example}
%io-programs of anthem can be implemented by using input predicates only in bodies, doing all the computation using only local predicates, finally reversing the result in the output predicate (with a check that it is not extended externally, as in tcg)

Finally, regarding \emph{c)}, the notions of \emph{relativised strong equivalence with projection}~\cite{DBLP:conf/ijcai/EiterTW05,DBLP:conf/jelia/GeibingerT19} and \emph{visible strong equivalence}~\cite{DBLP:journals/tocl/BomansonJN20} address the issue of excluding hidden predicates when verifying the invariant properties of (parts of) logic programs. 
These notions might provide material for extending the testing functionalities of our library beyond invariants based on plain here-and-there models.

\section{Conclusion}\label{sec:conclusion}

Templates introduce a naming convention to separate local and global names, and transpilation to ordinary ASP so to map local names to universally unique predicates.
This way transpiled programs can be simply combined by concatenation with the invariant that local names of different template applications do not clash.
Such an invariant can enforce other invariants, as for example ensuring that a global predicate is not further extended by other rules, including those that have not been written yet.
Some testing functionalities in this direction are given in Section~\ref{sec:properties}, and more are expected in our future work.
Finally, we expect to enrich the core template library with other common patterns of ASP.

\bibliographystyle{splncs04}
\bibliography{bibtex}

\begin{thebibliography}{10}
\providecommand{\url}[1]{\texttt{#1}}
\providecommand{\urlprefix}{URL }
\providecommand{\doi}[1]{https://doi.org/#1}

\bibitem{DBLP:journals/isse/AlOmarWRMNO22}
AlOmar, E.A., Wang, T., Raut, V., Mkaouer, M.W., Newman, C.D., Ouni, A.:
  Refactoring for reuse: an empirical study. Innov. Syst. Softw. Eng.
  \textbf{18}(1),  105--135 (2022)

\bibitem{DBLP:conf/jelia/AmendolaBR21}
Amendola, G., Berei, T., Ricca, F.: Testing in {ASP:} revisited language and
  programming environment. In: {JELIA}. Lecture Notes in Computer Science, vol.
  12678, pp. 362--376. Springer (2021)

\bibitem{DBLP:conf/iclp/BaralDT06}
Baral, C., Dzifcak, J., Takahashi, H.: Macros, macro calls and use of ensembles
  in modular answer set programming. In: {ICLP}. Lecture Notes in Computer
  Science, vol.~4079, pp. 376--390. Springer (2006)

\bibitem{DBLP:journals/tocl/BomansonJN20}
Bomanson, J., Janhunen, T., Niemel{\"{a}}, I.: Applying visible strong
  equivalence in answer-set program transformations. {ACM} Trans. Comput. Log.
  \textbf{21}(4),  33:1--33:41 (2020)

\bibitem{DBLP:journals/tplp/CabalarFL20}
Cabalar, P., Fandinno, J., Lierler, Y.: Modular answer set programming as a
  formal specification language. Theory Pract. Log. Program.  \textbf{20}(5),
  767--782 (2020)

\bibitem{DBLP:journals/tplp/CalimeriFGIKKLM20}
Calimeri, F., Faber, W., Gebser, M., Ianni, G., Kaminski, R., Krennwallner, T.,
  Leone, N., Maratea, M., Ricca, F., Schaub, T.: Asp-core-2 input language
  format. Theory Pract. Log. Program.  \textbf{20}(2),  294--309 (2020)

\bibitem{DBLP:journals/aicom/CalimeriI06}
Calimeri, F., Ianni, G.: Template programs for disjunctive logic programming:
  An operational semantics. {AI} Commun.  \textbf{19}(3),  193--206 (2006)

\bibitem{DBLP:conf/iclp/Dao-TranEFK09}
Dao{-}Tran, M., Eiter, T., Fink, M., Krennwallner, T.: Modular nonmonotonic
  logic programming revisited. In: {ICLP}. Lecture Notes in Computer Science,
  vol.~5649, pp. 145--159. Springer (2009)

\bibitem{DBLP:conf/kr/Dao-TranEFK10}
Dao{-}Tran, M., Eiter, T., Fink, M., Krennwallner, T.: Distributed nonmonotonic
  multi-context systems. In: {KR}. {AAAI} Press (2010)

\bibitem{DBLP:conf/lpnmr/EiterGV97}
Eiter, T., Gottlob, G., Veith, H.: Modular logic programming and generalized
  quantifiers. In: {LPNMR}. Lecture Notes in Computer Science, vol.~1265, pp.
  290--309. Springer (1997)

\bibitem{DBLP:conf/ijcai/EiterTW05}
Eiter, T., Tompits, H., Woltran, S.: On solution correspondences in answer-set
  programming. In: {IJCAI}. pp. 97--102. Professional Book Center (2005)

\bibitem{DBLP:journals/aim/ErdemGL16}
Erdem, E., Gelfond, M., Leone, N.: Applications of answer set programming. {AI}
  Mag.  \textbf{37}(3),  53--68 (2016)

\bibitem{DBLP:journals/tplp/FandinnoLLS20}
Fandinno, J., Lifschitz, V., L{\"{u}}hne, P., Schaub, T.: Verifying tight logic
  programs with anthem and vampire. Theory Pract. Log. Program.
  \textbf{20}(5),  735--750 (2020)

\bibitem{DBLP:journals/algorithms/FandinnoPVW22}
Fandinno, J., Pearce, D., Vidal, C., Woltran, S.: Comparing the reasoning
  capabilities of equilibrium theories and answer set programs. Algorithms
  \textbf{15}(6), ~201 (2022)

\bibitem{DBLP:conf/cilc/FebbraroRR11}
Febbraro, O., Reale, K., Ricca, F.: Testing {ASP} programs in {ASPIDE}. In:
  {CILC}. {CEUR} Workshop Proceedings, vol.~810, pp. 115--129. CEUR-WS.org
  (2011)

\bibitem{DBLP:journals/tplp/Fink11}
Fink, M.: A general framework for equivalences in answer-set programming by
  countermodels in the logic of here-and-there. Theory Pract. Log. Program.
  \textbf{11}(2-3),  171--202 (2011)

\bibitem{DBLP:conf/jelia/GeibingerT19}
Geibinger, T., Tompits, H.: Characterising relativised strong equivalence with
  projection for non-ground answer-set programs. In: {JELIA}. Lecture Notes in
  Computer Science, vol. 11468, pp. 542--558. Springer (2019)

\bibitem{DBLP:conf/lpnmr/GresslerOT17}
Gre{\ss}ler, A., Oetsch, J., Tompits, H.: Harvey: {A} system for random testing
  in {ASP}. In: {LPNMR}. Lecture Notes in Computer Science, vol. 10377, pp.
  229--235. Springer (2017)

\bibitem{HeytingDieFR}
Heyting, A.: Die formalen regeln der intuitionistischen logik. pp. 42--56.
  Deütsche Akademie der Wissenschaften zu Berlin,
  Mathematisch-Naturwissenschaftliche Klasse (1930)

\bibitem{DBLP:conf/ecai/JanhunenNOPT10}
Janhunen, T., Niemel{\"{a}}, I., Oetsch, J., P{\"{u}}hrer, J., Tompits, H.: On
  testing answer-set programs. In: {ECAI}. Frontiers in Artificial Intelligence
  and Applications, vol.~215, pp. 951--956. {IOS} Press (2010)

\bibitem{DBLP:conf/lpnmr/JanhunenNOPT11}
Janhunen, T., Niemel{\"{a}}, I., Oetsch, J., P{\"{u}}hrer, J., Tompits, H.:
  Random vs. structure-based testing of answer-set programs: An experimental
  comparison. In: {LPNMR}. Lecture Notes in Computer Science, vol.~6645, pp.
  242--247. Springer (2011)

\bibitem{DBLP:journals/jair/JanhunenOTW09}
Janhunen, T., Oikarinen, E., Tompits, H., Woltran, S.: Modularity aspects of
  disjunctive stable models. J. Artif. Intell. Res.  \textbf{35},  813--857
  (2009)

\bibitem{DBLP:journals/tplp/KaminskiRSW23}
Kaminski, R., Romero, J., Schaub, T., Wanko, P.: How to build your own
  asp-based system?! Theory Pract. Log. Program.  \textbf{23}(1),  299--361
  (2023)

\bibitem{RFC4122}
Leach, P., Mealling, M., Salz, R.: A universally unique identifier (uuid) urn
  namespace. Internet Requests for Comments (July 2005),
  \url{https://tools.ietf.org/html/rfc4122}

\bibitem{DBLP:journals/tplp/Lifschitz17}
Lifschitz, V.: Achievements in answer set programming. Theory Pract. Log.
  Program.  \textbf{17}(5-6),  961--973 (2017)

\bibitem{DBLP:books/sp/Lifschitz19}
Lifschitz, V.: Answer Set Programming. Springer (2019)

\bibitem{DBLP:journals/corr/abs-1905-03196}
L{\"{u}}hne, P.: Discovering and proving invariants in answer set programming
  and planning. CoRR  \textbf{abs/1905.03196} (2019)

\bibitem{DBLP:conf/kr/OetschPPST12}
Oetsch, J., Prischink, M., P{\"{u}}hrer, J., Schwengerer, M., Tompits, H.: On
  the small-scope hypothesis for testing answer-set programs. In: {KR}. {AAAI}
  Press (2012)

\bibitem{DBLP:journals/tocl/OetschSTW21}
Oetsch, J., Seidl, M., Tompits, H., Woltran, S.: Beyond uniform equivalence
  between answer-set programs. {ACM} Trans. Comput. Log.  \textbf{22}(1),
  2:1--2:46 (2021)

\bibitem{DBLP:journals/amai/Pearce06}
Pearce, D.: Equilibrium logic. Ann. Math. Artif. Intell.  \textbf{47}(1-2),
  3--41 (2006)

\bibitem{DBLP:journals/tplp/SonPBS23}
Son, T.C., Pontelli, E., Balduccini, M., Schaub, T.: Answer set planning: {A}
  survey. Theory Pract. Log. Program.  \textbf{23}(1),  226--298 (2023)

\bibitem{DBLP:journals/tplp/VosKOPT12}
Vos, M.D., Kisa, D.G., Oetsch, J., P{\"{u}}hrer, J., Tompits, H.: Annotating
  answer-set programs in lana. Theory Pract. Log. Program.  \textbf{12}(4-5),
  619--637 (2012)

\end{thebibliography}

\clearpage
\appendix

\section{Proofs}

\begin{proof}[Proposition~\ref{prop:models}]
$\mathit{HT}(\Gamma_\Pi) \supseteq \mathit{HT}(\Gamma_\Pi \cup \Gamma_{\Pi'})$ follows from the monotonicity of here-and-there, and in turn implies \eqref{eq:true-inv}--\eqref{eq:reduct-inv}.
\hfill\qed
\end{proof}

\begin{proof}[Proposition~\ref{prop:unstable}]
The only interesting case is $I \cup X \models \Pi \cup \Pi'$, for which we shall show that
$I \models (\Pi \cup \Pi')^{I \cup X}$.
Since $\tuple{I,I \cup X} \in \mathit{HT}(\Gamma_{\Pi})$ by assumption, we thus have \linebreak
$I \models \Pi^{I \cup X}$, and therefore it remains to show 
$I \models (\Pi')^{I \cup X}$.
From the assumption $I \cup X \models \Pi \cup \Pi'$, we have $I \cup X \models \Pi'$;
combining with the assumption that atoms in $X$ have predicates in $\mathit{pred}(\Pi) \setminus \mathit{pred}(\Pi')$, we can conclude that $I \models \Pi'$ and $(\Pi')^{I \cup X} = (\Pi')^I$.
Hence, $I \models (\Pi')^{I \cup X}$ and we are done.
\hfill\qed
\end{proof}

\begin{proof}[Theorem~\ref{thm:local}]
By construction, $\mathit{pred}(\pi\rho) \cap \mathbf{P_L}$ are universally unique, and therefore they cannot occur in $\Pi'$.
\hfill\qed
\end{proof}

\end{document}